# Developing Universal Dependency Treebanks for Magahi and Braj

**Mohit Raj**[!], **Shyam Ratan**[!], **Deepak Alok**[+],
**Ritesh Kumar**[!], **Atul Kr. Ojha**[*,+]

[!] Dr. Bhimrao Ambedkar University, Agra  [+] Panlingua Language Processing LLP, New Delhi
[*] Data Science Institute, National University of Ireland Galway, Ireland

mohiitraj@gmail.com   shyamratan2907@gmail.com   deepak06alok@gmail.com
riteshkr.kmi@gmail.com   panlingua@outlook.com

## Abstract

In this paper, we discuss the development of treebanks for two low-resourced Indian languages - Magahi and Braj - based on the Universal Dependencies framework. The Magahi treebank contains 945 sentences and Braj treebank around 500 sentences marked with their lemmas, part-of-speech, morphological features and universal dependencies. This paper gives a description of the different dependency relationship found in the two languages and give some statistics of the two treebanks. The dataset will be made publicly available on Universal Dependency (UD) repository (https://github.com/UniversalDependencies/UD_Magahi-MGTB/tree/master) in the next (v2.10) release.

## 1 Introduction

Magahi is an Eastern Indo-Aryan Language, spoken mainly in Eastern Indian states including Bihar and Jharkhand, along with some parts of West Bengal and Odisha. Magahi is classified under the Eastern group of the outer sub-branch of Indo-Aryan language (Grierson, 1908). Scholars like Turner have clubbed the 'Bihari' languages with Eastern and Western Hindi (Masica, 1991). There is another kind of classfication of Indian language in which western Hindi is almost an isolated group while Eastern Hindi, Bihari and other languages of Eastern group are clubbed together (Chatterji, 1926). But the classification in which Magahi comes under the Eastern group of the outer sub-branch of Indo-Aryan language is the most widely accepted classification.

Brajbhasha is classified in two ways. According to first classification, Brajbhasha is a Western Indo-Aryan language that is spoken in the states of Western Uttar Pradesh and parts of Rajasthan (Jeffers, 1976). The other classification puts Brajbhasha in the group of Western Hindi of Central Group of Indo-Aryan sub-family of Indo European language family along with Hindustani, Bangaru, Brajbhaka, Kanauji, Bundeli (Grierson, 1908).

The difficulty of tracing the exact historical path of a large number of these Indo-Aryan languages is discussed in detail by Masica (Masica, 1991) and is evident by somewhat incompatible classification given by Chatterji, Turner, Katre, Cardona and Mitra and several other scholar (Masica, 1991);(Chatterji, 1926); (Turner, 1966); (Katre, 1968); (Cardona, 1974); (Mitra et al., 1978). As such the exact status of Magahi and Braj vis-a-vis other Indo-Aryan languages (especially the major ones like Hindi, Bangla and Odia) remains hazy and controversial. This, coupled with the imposition of Modern Standard Hindi (MSH) over what is now popularly known as 'Hindi Belt' and what has historically been established as a rather complex dialect continuum, with several languages and varieties being spoken in different domains of usage (Gumperz, 1957), has resulted in these languages mistakenly classified as varieties of Hindi. This, in turn, has resulted in not only minimal support from the Government for development of different kinds of resources for the language but also a negative attitude of the speakers towards the language (Kumar et al., 2018a).

However, despite this disadvantageous situation of the language, there has been some efforts at developing language technologies and resources for these languages, especially for Magahi viz. monolingual written and speech corpora (Kumar et al., 2014), Magahi part-of-speech tagger (Kumar et al., 2012) and Magahi language identification system (Rani et al., 2018). This paper describes another such effort towards developing a Universal Dependency (UD) based treebank for the language which may prove to be useful in processing and analysing the language for different applications.

## 2 Universal Dependencies and Low-resource Languages

Universal Dependencies framework provides some unique advantages for low-resource languages both in terms of making the language for cross-lingual comparison and studies as well as making transfer learning and multilingual techniques for technology development possible. As a result in recent times we have seen the development of UD treebanks for quite a few low-resource languages viz. Yorùbá (Ishola and Zeman, 2020), Latin Treebank for UD (Cecchini et al., 2020), Hittite (Andersen and Rozonoyer, 2020), Manx Gaelic (Scannell, 2020), Laz (Türk et al., 2020), Albanian (Toska et al., 2020) and others.

The UD treebanks have also been built for Indian languages such as Bhojpuri, Hindi, Marathi, Sanskrit, Tamil, Telugu and Urdu (Zeman and et al., 2021; Ojha and Zeman, 2020). Except Hindi, all of the (Indian) languages mentioned above are low-resourced languages. Recently, Dash et al. (2021) reported the development of a treebank in Santhali, another low-resourced language spoken in India.

However, one of the biggest challenges in building treebanks for a large number of low-resource languages is the absence of grammatical descriptions and hence a reference point for deciding on the analysis needed to give the dependency relationships. There are many treebanks that could be roughly classified in two groups. Treebanks of well-known and well-described languages like Hindi, English, French etc and treebanks of lesser known and sparsely described languages. Magahi and Braj come in the second group which are sparsely described languages. There have been very few linguistic studies on these languages with both of these languages lacking an exhaustive grammatical description or even a dictionary. These are some linguistics studies towards the description of Magahi: a basic (although not completely accurate) description of Magahi is given by Shila Verma (Verma and Verma, 1983; Verma, 1985), a description of Magahi case system (Lahiri, 2021, 2014; Kumar et al., 2014), discussion on Magahi honorific system within the minimalist framework (Alok, 2021), morphosyntactic properties of nominal particle -wa (Alok, 2014), and study on the Magahi spatial postpositions (Alok, 2012).

For Braj, to the best of our knowledge, the only modern linguistic studies are on its ergativity under the minimalist framework (Chandra and Kaur, 2020a,b). .

This lack of an exhaustive description of different aspects of Magahi and Braj morphosyntax made the task of developing the treebank quite challenging and required establishing multiple grammatical analyses of the languages while working on the treebank. The aim of this paper is to give a broad description of Magahi and Braj morphosyntax within the Universal Dependencies framework with respect to different syntactic dependency relationships in the languages, along with a discussion on the process of the development of this treebank.

## 3 Treebank Creation in Magahi and Braj

We annotate the Magahi and Braj treebank with lemma, Universal Parts-of-Speech (POS) tags, a subset of morphological features and the Universal dependency relation. The dependency relation for Magahi and Braj are annotated using a subset of the 37 dependency relations included in the Universal Dependency tagset[1]. In the following sections, we describe each of the features marked for building Magahi and Braj dependency treebank. Most of these relationships follow the canonical patterns as discussed in the UD guidelines (and as witnessed in common Indo-European languages).

---

[1] https://universaldependencies.org/u/dep/

### 3.1 POS and Morphological Features

We use the Universal POS tags for annotating the POS tags for the data. However, for morphological features we mark only those features which have explicit morphological realisation in the two languages. Thus, for example, we mark gender and number on Braj verbs but not on Magahi verbs since there is no number or gender agreement in Magahi. Table 1 gives a list of all the morphological features and their values that we mark for each category of words in each of the language [2].

### 3.2 Core Dependency arguments

#### 3.2.1 Nominal Subject : nsubj

Nominal subject in Magahi and Braj plays the role of syntactic subject and it is dependent on the verb. Let us take a look at the Figure 1 and 2.

- राजा लाल लेके रानी के दे देलन ।

  raɟa: lal leke rani: ke ɖe ɖelən .

  King took the perl and gave to the queen.

- मैं एक गाँव की पाठशाला में कक्षा दूसरी कूँ पढ़ा रह्यौ ।

  mainⁿ ek gaⁿv ki: paʈʰʃala: meⁿ kəkʂa: ɖus-ri: ku:ⁿ pəɖʰə̞a: rəfija:u .

  I was teaching class second in a village school.

#### 3.2.2 Object : obj

Object of a sentence in the two languages is also dependent on the verb. Let us take a look at the Figure 1 and 2.

#### 3.2.3 Indirect Object : iobj

The role and nature of indirect objects is similar to direct objects except the syntactic closeness with verbs. Indirect objects are also dependent on verbs and this relationship is indicated by iobj. Let us take a look at the Figure 1 and 2.

---

²In the table * indicate that the given feature or value is marked only for Braj

#### 3.2.4 Clausal complement : ccomp

Clausal complement occurs with complex structure of sentence in Magahi and Braj. When the sentence is formed with two clauses (principle and subordinate), the root (generally, verb) of the subordinate clause depends on the root of the principal clause. Subordinate clause behaves like an object of the main clause.

#### 3.2.5 Open clausal complement : xcomp

Open clausal complement differs from the clausal complement in that in this case the head of the subordinate clause does not seem to have an overt subject and as such there is a dependence relation between the root of the subordinate clause and a word of the higher clause of Magahi and Braj.

### 3.3 Non-core dependents

#### 3.3.1 Oblique Nominal : obl

Oblique is the nominal element of a sentence which appears as an adjunctive argument and it depends upon the main verb of the sentence. Sometimes it adds extra information about a verb, adjective and adverb so it functionally acts as an adverbial attachment. Oblique is grammatically categorised as a noun or pronoun and used as a temporal and nominal locational modifier as in the given example of Magahi and passive agent are also labelled as oblique. In another example from Braj, the adverbial modifier also bears an obl relation with the verb.

#### 3.3.2 Adverbial clause modifier : advcl

Adverbial clause modifier is an entity of complex structural sentences in which it is the main predicate of a dependent clause. Like an adverb, its functional role is to modify a verb or other predicate such as an adjective of principal clause or other clausal entity, but the difference lies in the fact that adverbial clause modifier establishes an interclausal relationship.

#### 3.3.3 Adverbial modifier: advmod

Adverbial modifier defines the relationship in between main verb of a clause and the adverb. In the given example of Magahi word ਖੂਬ'/'kʰubə' is modifying the main verb or root ਨੱਚਬਵਡ'/'nəcəbəvə' of sentence. Another kind of adverbial modifier is given in the Braj example which is negating the act of event. Let us take a look at the Figure 4 and 3.

| Feature | UPOS | Values |
|---------|------|--------|
| Case | NOUN, PRON, ADP | Nom, Acc, Dat, Gen, Erg*, Abl |
| Gender* | NOUN, PRON, VERB, PROPN, | Fem, Masc |
| Number* | NOUN, PRON, VERB | Sing, Plur |
| Person | PRON, VERB | 1,2,3 |
| Tense | VERB, AUX | Pres, Past, Fut* |
| Aspect | VERB, AUX | Prog, Imp, Perf, Hab |
| Politeness | PRON, VERB, AUX | Form, Infm |

Table 1: Feature values for Magahi & Braj grouped by UPOS

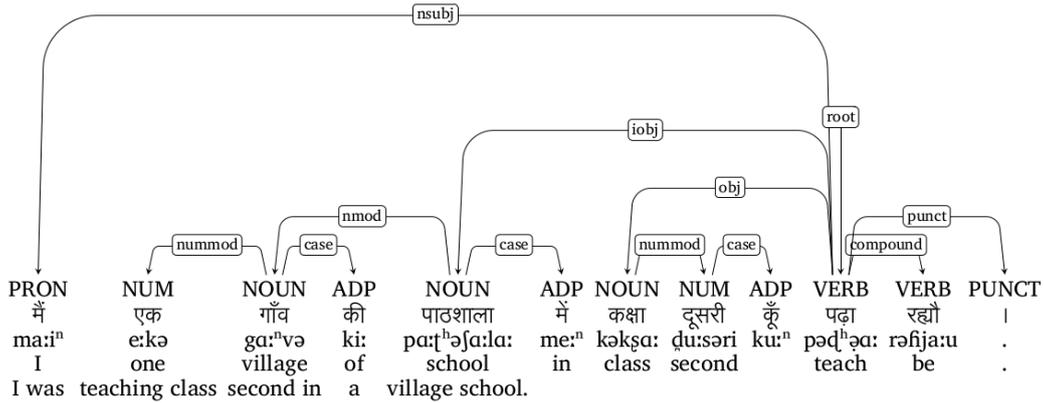

Figure 1: Braj Example for nsubj, obj, iobj

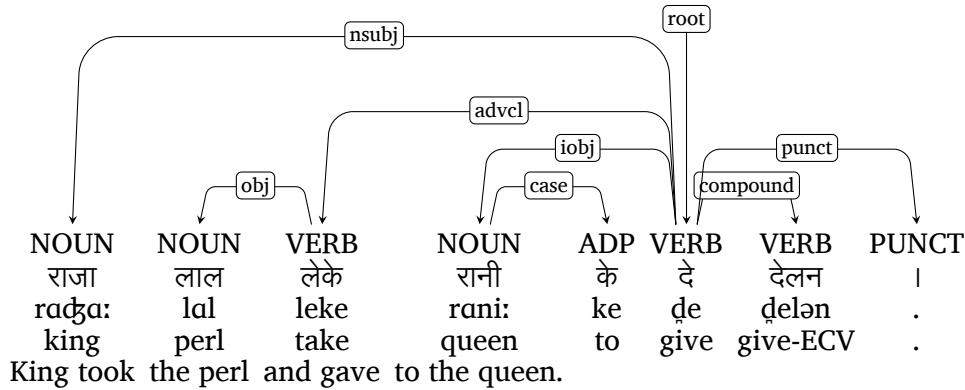

Figure 2: Magahi Example for nsubj, iobj, obj

- आउ हम खुब नचबवऽ ।

  ɑːu ɦəm kʰub nəcbəvə .

  And I will dance a lot.

- पै बूआ कहीं नाँय जाती।

  pai buɑː kəɦiːⁿ nɑːⁿj ɟɑːʈiː .

But the aunt did not go anywhere.

### 3.3.4 Auxiliary : aux

aux is the dependency relation between a verbal predicate and the auxiliary in the two languages. Let us take a look at the Figure 5.

- या कारन मैं चुप्प खींच गयौ हो ।

  jɑː kɑːrən maiⁿ cuppə kʰiːⁿc gəjɑːu ɦo .

  For this reason I remained silent.

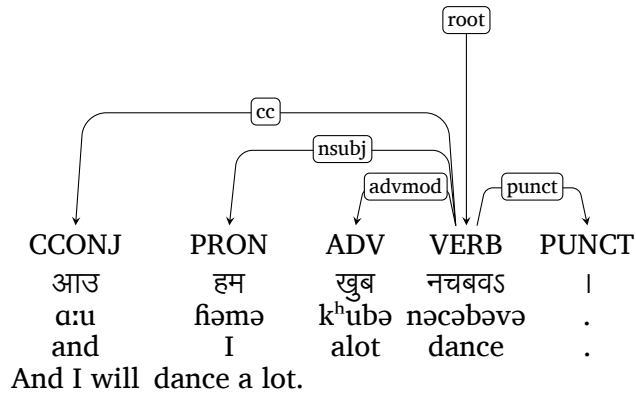

Figure 3: Magahi Example for advmod

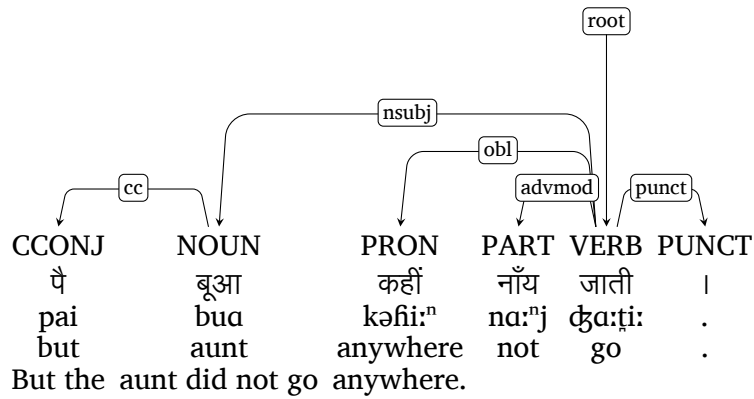

Figure 4: Braj Example for advmod

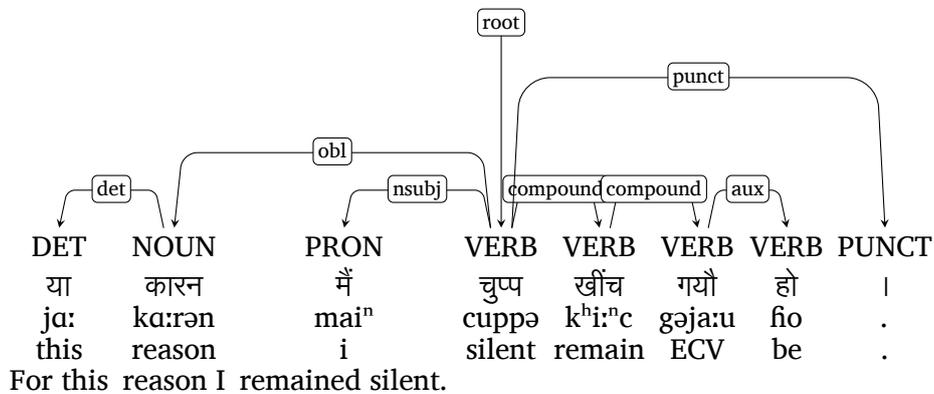

Figure 5: Braj Example for aux

### 3.3.5 Copula : cop

aux is the dependency relation between a non-verbal predicate and the auxiliary in the two languages. Let us take a look at the Figure 9 and 6.

- सीर्सक हौ " गोबिंद बढई " ।

  siːrsək ɦau " gobiⁿd̪ə bəɖʰəiː " .

Title was Govind carpenter.

We found two type of copula in Braj (simple and complex). Simple copula (हतौ/ɦəʈɒː, है/ɦai, ही/ɦiː) are like copula of Magahi, and complex copula (नांओ'/'naːⁿɒ', आओ'/'aːɒ', गओ'/'gəɒ') are not present in Magahi.

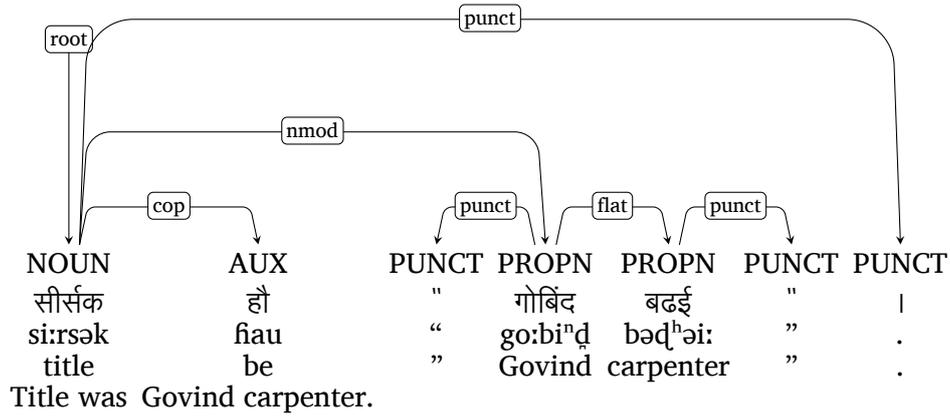

Figure 6: Braj Example for cop

### 3.3.6 Marker : mark

Typically the subordinating conjunction depends on the head of the subordinate clause - such dependency relationships are called mark.

## 3.4 Nominal dependents

### 3.4.1 Nominal modifier : nmod

The nominal modifier is noun or pronoun and it is syntactically dependent upon another noun or noun phrase and functionally corresponds to an attribute or genitive complement.

### 3.4.2 Numeral modifier : nummod

It is the relationship between a numeral and the noun which is attached to the numeral. Let us take a look at the Figure 1.

### 3.4.3 Clausal modifier of noun : acl

In the acl syntactic relation head is noun that is modified and the dependent is the head of the clause that modifies the noun.

### 3.4.4 Adjectival modifier : amod

amod defines the dependency relationship between the noun and the adjective.

### 3.4.5 Determiner : det

The relation between determiner and nominal head is annotate with syntactic relation det. Let us take a look at the Figure 7.

- मैं जा बहली में बैठ बू पलट गई ।

  maiⁿ ʤa: bəɦəli: meⁿ baiʈʰ bu: pələʈ gəi: .

  The bullock cart I was sitting in overturned.

### 3.4.6 Classifier : clf

Numeral classifiers are used only in Magahi and clf defines the relation between the numeral and the classifier. Let us take a look at the Figure 8.

- इधर राजा सात गो बियाह कर लेलन ।

  iḍʰərə raʤa: sa:ʈ go bija:ɦ kər lelən .

  Meanwhile king has married to seven girl.

### 3.4.7 Case marker : case

Postpositions are the case marking elements in both Braj and Magahi. The postpositions are syntactically dependent on the noun to which they attach and are related by the 'case' relation.

## 3.5 Coordination

### 3.5.1 Conjunct : conj

In case of coordinating conjunction construction, the head of the second clause depends upon the head of the first clause and are related by the 'conj' relation.

### 3.5.2 Coordinating Conjunction : cc

The coordinating conjunction itself is dependent upon the head word of the second clause or phrase and have a 'cc' relation. Let us take a look at the Figure 10 and 9.

- जैन बड़ी मजबुत आउ बहादुर हल ।

  ʤain bəɖəi: məʤbuʈ a:u bəɦa:ḍur ɦəl .

  Jain was very strong and brave.

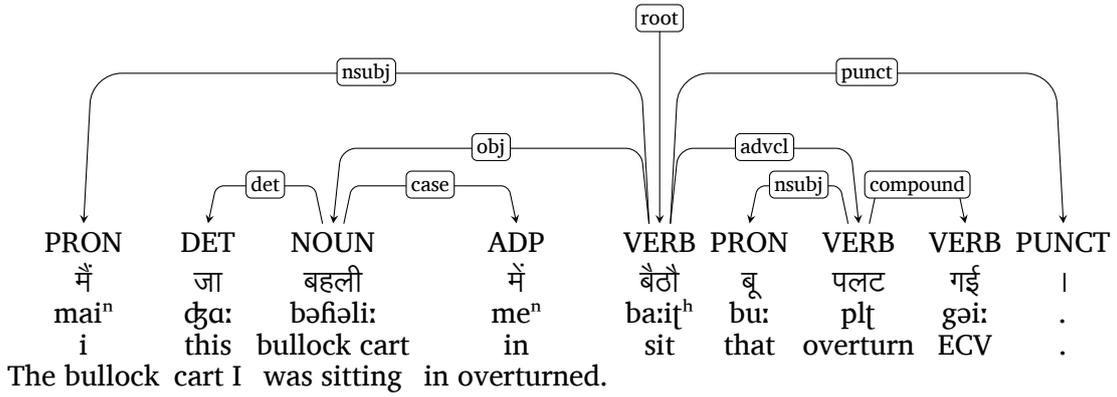

Figure 7: Braj Example for det

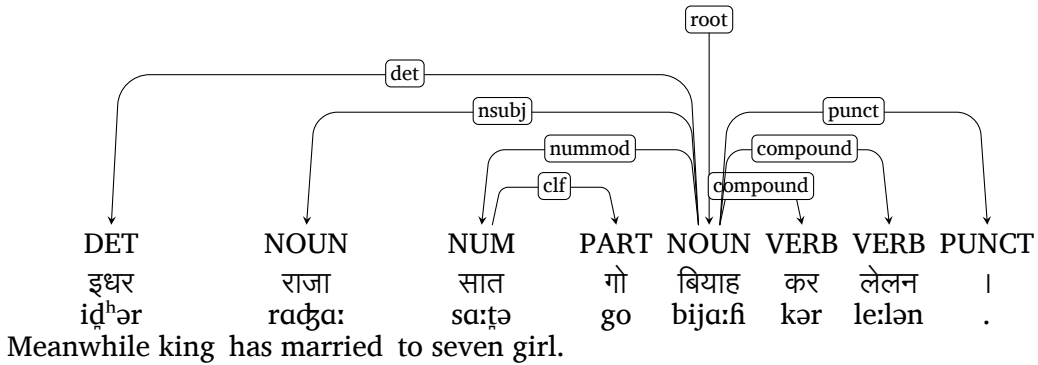

Figure 8: Magahi Example for clf

- झारू बुहारी कपड़ा – लत्ता और रसौई ।
- dʒʰɑ:ru: buɦɑ:ri: kəpəɖə̣ɑ: - ləṭṭa: aur rəsaui: .

  Broom, cloth and kitchen.

### 3.6 Multi Word Expressions

#### 3.6.1 Flat multiword expression : flat

The flat is a syntactic relation that is used for such multiword expression in which there is no specific head word like in name(Sohan Lal Mishra) or dates(2nd B.C.) etc.

#### 3.6.2 compound : compound

A compound syntactic relation is used for different kinds of multiword expressions in Magahi and Braj including compound nouns, compound verbs, conjunct verbs (noun/adjective + verb), echo words, and reduplication. While most of these constructions are quite predominant in almost all of the South Asian languages and could actually have different kinds of syntactic and semantic impacts, unfortunately UD provides very limited ways of distinguishing across these and it proved to be one of the most challenging aspects of building this treebank.

## 4 Magahi and Braj Treebank

We have used Magahi and Braj plain text to prepare a treebank. Magahi plain text is part of a large monolingual written and speech corpora (Kumar et al., 2014), which is prepared from the Magahi literature. Braj plain text is also prepared from the literary domain (Kumar et al., 2018b). We have used Conllueditor (Heinecke, 2019) tool to build a treebank. The tool facilitate us to attach several kinds of information with words like UPOS, lemma, morph feature and dependency relation among different word of sentence etc.

Currently, Braj treebank has a total of around 5.8k tokens (excluding punctuations) in a total of 500 sentences. Magahi, on the other hand, has a total of over 12k tokens from 945 sentences. Table 2 and Table 3 gives the detailed statistics of each UPOS and UD category and morphological features in the two treebanks. As expected, nouns and verbs form the most predominant POS categories in both the languages, both of them together ac-counting for over 40% of the tokens. These are followed by adpositions and pronouns as the most frequent category of words in the treebank.

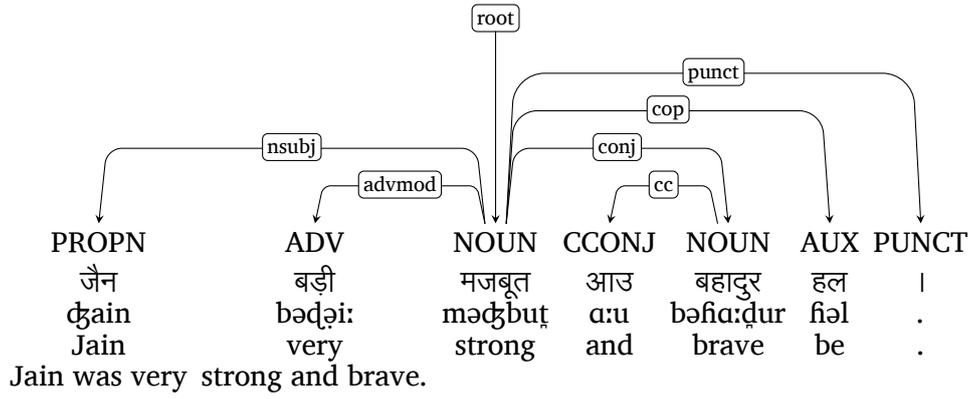

Figure 9: Magahi Example for cc and cop

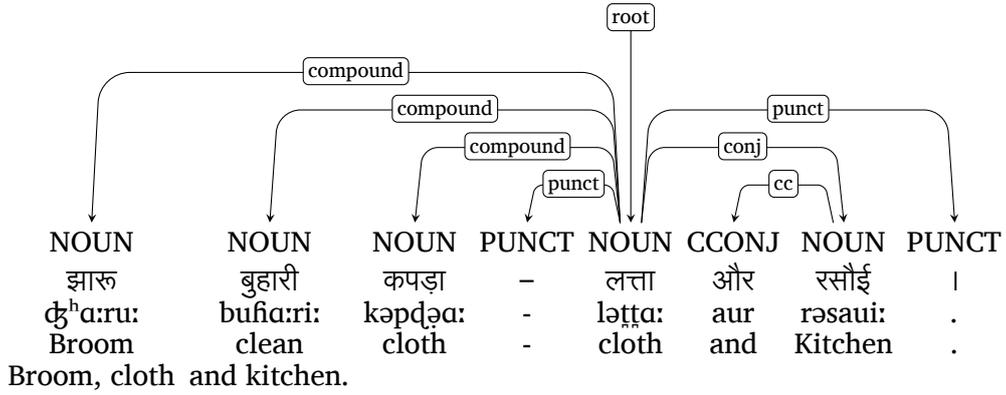

Figure 10: Braj Example for cc

| UPOS | Braj Count | Braj % | Magahi Count | Magahi % |
|---|---|---|---|---|
| NOUN | 1507 | 23.17 % | 3203 | 24.01 % |
| VERB | 1203 | 18.49 % | 3040 | 22.78 % |
| PART | 224 | 3.44 % | 282 | 2.11 % |
| PRON | 562 | 8.64 % | 1172 | 8.78 % |
| CCONJ | 113 | 1.73 % | 330 | 2.47 % |
| ADV | 83 | 1.27 % | 273 | 2.05 % |
| PROPN | 246 | 3.78 % | 249 | 1.87 % |
| ADP | 778 | 11.96 % | 1681 | 12.60 % |
| SCONJ | 112 | 1.72 % | 418 | 3.13 % |
| NUM | 253 | 3.89 % | 356 | 2.67 % |
| ADJ | 276 | 4.24 % | 168 | 1.26 % |
| DET | 181 | 2.78 % | 385 | 2.89 % |
| AUX | 230 | 3.53 % | 440 | 3.30 % |
| INTJ | 0 | 0 % | 15 | 0.11 % |
| PUNCT | 842 | 12.94 % | 1331 | 9.98 % |
| **TOTAL** | **6,610** | **100 %** | **13,343** | **100 %** |

Table 2: Braj & Magahi UPOS Category Statistics

| Dependency Relation | Braj Count | Braj % | Magahi Count | Magahi % |
|---|---|---|---|---|
| root | 500 | 7.41 % | 945 | 7.01% |
| obj | 444 | 6.58 % | 808 | 5.99% |
| nmod | 547 | 8.11 % | 531 | 3.94% |
| nsubj | 578 | 8.57 % | 1065 | 7.90% |
| obl | 106 | 1.57 % | 580 | 4.30% |
| advmod | 169 | 2.50 % | 419 | 3.11% |
| cc | 113 | 1.67 % | 313 | 2.32% |
| case | 771 | 11.44 % | 1861 | 13.81% |
| conj | 516 | 7.65 % | 320 | 2.37% |
| mark | 126 | 1.86 % | 419 | 3.11% |
| advcl | 244 | 3.62 % | 748 | 5.55% |
| nummod | 117 | 1.73 % | 293 | 2.17% |
| iobj | 188 | 2.78 % | 851 | 6.31% |
| det | 172 | 2.55 % | 338 | 2.50% |
| amod | 211 | 3.11 % | 166 | 1.23% |
| xcomp | 28 | 0.41 % | 76 | 0.56% |
| aux | 146 | 2.16 % | 296 | 2.19% |
| cop | 96 | 1.42 % | 126 | 0.93% |
| compound | 577 | 8.56 % | 1327 | 9.85% |
| dep | 99 | 1.46 % | 163 | 1.20% |
| ccomp | 13 | 0.19 % | 312 | 2.31% |
| acl | 3 | 0.044 % | 54 | 0.40% |
| flat | 122 | 1.81 % | 124 | 0.92% |
| clf | 0 | 0 % | 6 | 0.04% |
| punct | 842 | 12.49 % | 1331 | 9.87% |

Table 3: Braj & Magahi Dependency Relation Statistics

## 5 Conclusion

In this paper, we have discussed the development of treebanks for Braj and Magahi - two extremely low-resource Eastern Indo-Aryan languages spoken in India. The treebank is annotated with lemma, UPOS, morphological features and UD relations. As of now the Braj treebank has 500 sentences (with around 5.8k tokens) while the Magahi treebank has 945 sentences (with over 12k tokens). A comparative analysis of dependency relation of Magahi and Braj treebank reveal that, as expected, most of the syntactic relations are shared across the two languages except the two dependency relation in our available dataset. The first one is copula (cop) relation - Magahi has a simple syntactic structure of cop while in Braj it could have a complex structure and it may vary between two types of lexical structure. The second dependency relation is that of classifier (clf) - Magahi has numeral classifier while this is not present in Braj.

## 6 Future Work

The analysis of the two languages as well as the development of the treebank is currently in progress - we are exploring the semi-automatic means of further increasing the treebank size such as developing and using parsers for annotating the data and then manually validating it. We are also exploring ways of getting data from varied domains (including narrations and conversational data) for including in the treebank.

## Acknowledgments

We would like to express our heartfelt gratitude to Panlingua Language Processing LLP for supporting the creation of these treebanks both academically and financially. We would also like to thank Dr. Mayank for helping us out in figuring out certain aspects of the Braj grammar and annotation.

Atul Kr. Ojha would like to acknowledge the EU's Horizon 2020 Research and Innovation programme through the ELEXIS project under grant agreement No. 731015.